\documentclass{article}

\PassOptionsToPackage{numbers, compress}{natbib}
\usepackage[education,final]{neurips_2026}
\makeatletter
\renewcommand{\@noticestring}{}
\makeatother

\usepackage[utf8]{inputenc}
\usepackage[T1]{fontenc}
\usepackage{hyperref}
\usepackage{url}
\usepackage{booktabs}
\usepackage{amsfonts}
\usepackage{nicefrac}
\usepackage{microtype}
\usepackage{xcolor}
\usepackage{graphicx}
\usepackage{tcolorbox}
\usepackage{enumitem}
\usepackage{float}
\usepackage{amsmath}

\setlist{noitemsep, topsep=2pt, parsep=0pt, partopsep=0pt, leftmargin=1.4em}
\hypersetup{colorlinks, linkcolor=blue} 
\makeatletter
\renewenvironment{table}{\@float{table}}{\end@float}
\makeatother

\usepackage{capt-of}

\usepackage{wrapfig}
\usepackage{needspace}

\newcommand{\minidogicon}{\raisebox{-0.22em}{\includegraphics[height=1.05em]{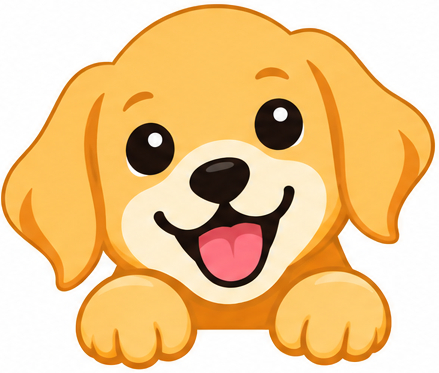}}}
\newcommand{\snowemoji}{\raisebox{-0.15em}{\includegraphics[height=0.95em]{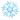}}}
\newcommand{\fireemoji}{\raisebox{-0.15em}{\includegraphics[height=0.95em]{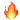}}}
\definecolor{hilite}{HTML}{FF1493}
\definecolor{hilitepurple}{HTML}{9400D3}
\definecolor{hiliteorange}{HTML}{FF8C00}

\definecolor{tokText}{HTML}{F8D6C5}
\definecolor{tokTime}{HTML}{ECBAE7}
\definecolor{tokLatent}{HTML}{CAECBD}
\definecolor{tokDino}{HTML}{C6C6C6}
\newcommand{\tokensq}[1]{\raisebox{-0.1em}{\tikz\fill[#1, rounded corners=1pt] (0,0) rectangle (0.85em,0.85em);}}

\tcbuselibrary{skins}

\usetikzlibrary{arrows.meta}
\definecolor{takeawayframe}{HTML}{004C4C}
\definecolor{takeawayback}{HTML}{F2F2F2}

\title{\minidogicon\,MiniDog: 
Dog-Breed Text-to-image Flow Matching from Pretraining 
to Fine-Tuning on 4 Consumer GPUs}

\author{%
  Reyhaneh Esmailizadeh \quad Xingjian Leng \quad Zhanhao Liang \quad Liang Zheng \\
  The Australian National University,   {\small\texttt{\{firstname.lastname\}@anu.edu.au}}
 \\
}

\begin{document}
\raggedbottom

\maketitle

\begin{abstract}
We generally perceive that training text-to-image diffusion models requires heavy compute: it is true if we aim at a comprehensive frontier model. This understanding prevents people from training their own models when getting started in the field. We develop MiniDog, a framework that enables pretraining diffusion transformers in 55 minutes and fine-tuning in another 15 minutes on four 3090 GPUs. For data, we use 26k dog images from ImageNet and caption them for pretraining and use 2k images generated by FLUX.1-schnell for fine-tuning. We precompute the image and text features and cache them for efficiency. For architecture, we use E2E-INVAE for its semantic latent space, a small transformer with 22M parameters, a small Qwen model for text encoding, and in-context conditioning. For training, we use iREPA and an EUPE perception encoder for acceleration. For evaluation, our repository supports FID, HPSv2 score, PickScore, and various VAE reconstruction metrics. The project is open-sourced at \href{https://github.com/Reyhaneesmailizadeh/MiniDog}{GitHub}. We do not claim technical novelty.

\end{abstract}




\begin{figure}[h]
  \centering
  \includegraphics[width=\linewidth]{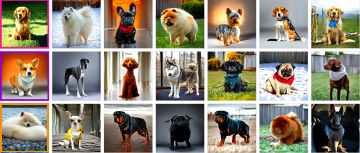}
  \caption{Samples from the SFT model. Prompts of the highlighted samples:
    \textcolor{hilite}{\rule{0.7em}{0.7em}}~``Golden Retriever with a dark golden, freshly brushed\dots''
    \textcolor{hilitepurple}{\rule{0.7em}{0.7em}}~``Corgi with a clean fawn-and-white coat stands alert\dots''
    \textcolor{hiliteorange}{\rule{0.7em}{0.7em}}~``Samoyed with a fluffy biscuit double coat sleeps curled\dots''}
    
  \label{fig:misc}
\end{figure}

\section{Introduction}

Training diffusion models is computationally heavy, especially for frontier models. In the community, beginners often struggle to find the opportunity to do the whole pipeline. This paper presents MiniDog, a simple yet useful codebase, where pretraining and fine-tuning of text-to-image (T2I) models can be performed in about an hour and 15 minutes, respectively, on four 3090 GPUs.

\textbf{Concept.} This material presents data, architecture, training methods, and evaluation metrics for T2I model training. Flow-matching beginners can train their own T2I models efficiently. 

\textbf{Leveling and prerequisite knowledge.} The material targets advanced undergraduate and graduate courses and assumes linear algebra, calculus, probability, and deep learning.

\textbf{Learning objectives and outcomes.} On completing the material, beginners can:

\begin{itemize}
  \item build and pretrain a text-to-image flow-matching pipeline on consumer GPUs (Section~\ref{sec:materials});
  \item explain how the tokenizer, caption length, and representation alignment change FID; and

  \item fine-tune the model and evaluate it with human-preference models.
\end{itemize}

\section{Background and related work}
\label{sec:related}

\paragraph{Flow matching and diffusion transformers.}
Diffusion models~\citep{ddpm} and flow matching~\citep{lipman2022flow} corrupt a clean latent $z_0$ with Gaussian noise $\epsilon\sim\mathcal{N}(0,I)$ in the same form, $z_t = \alpha_t z_0 + \sigma_t\epsilon$ for $t\in[0,1]$. They differ only in the noise schedule, which is how the weights $\alpha_t$ and $\sigma_t$ change with $t$. Diffusion models typically use $\alpha_t=\sqrt{\bar{\alpha}_t}$ and $\sigma_t=\sqrt{1-\bar{\alpha}_t}$, where $\bar{\alpha}_t$ follows a specific schedule, while rectified flow~\citep{rectifiedflow} takes the straight path $\alpha_t=1-t$, $\sigma_t=t$. We use rectified flow with this straight path throughout. The network $v_\theta$, conditioned on $\mathcal{F}_{\text{text}}$, predicts the path velocity $\mathrm{d}z_t/\mathrm{d}t=\epsilon-z_0$:
\begin{equation}
\mathcal{L}_{\text{FM}} = \mathbb{E}_{z_0,\epsilon,t}\,\bigl\lVert v_\theta(z_t, t, \mathcal{F}_{\text{text}}) - (\epsilon - z_0)\bigr\rVert^2 .
\label{eq:fm}
\end{equation}
In the sampling phase, noise at $t=1$ gets back to data at $t=0$ by following the velocity predicted by the denoising model. The denoiser is now typically a transformer. Diffusion Transformer (DiT)~\citep{peebles2023scalable} establishes this architecture by adapting Vision Transformers~\citep{vit} to latent space, which has replaced legacy U-Nets~\citep{unet}. We use LightningDiT~\citep{lightningdit}, a more efficient variant that improves the standard DiT backbone with training upgrades such as rectified flow and logit-normal timestep sampling~\citep{sd3}. It converges faster than the standard DiT without losing image quality.

\paragraph{Representation alignment} accelerates diffusion model training \citep{repa}. In REPA, the intermediate layers of the student model align to the output features of the teacher model, making the learning model rely both on its own generative capability and teacher knowledge. In other words, REPA takes the hidden states $h$ of the $N$ patch tokens at an intermediate block, maps each through a small learned projector $g_\phi$, and pulls it toward the target feature of the same patch of the clean image: 
\begin{equation}
\mathcal{L}_{\text{align}}(\theta,\phi) = \mathbb{E}_{t,z_0,\epsilon}\Bigl[\frac{1}{N}\sum_{n=1}^{N} d\bigl(g_\phi(h^{[n]}),\,\mathcal{F}_{\text{perception}}^{[n]}\bigr)\Bigr],
\qquad
\mathcal{L} = \mathcal{L}_{\text{FM}} + \lambda\,\mathcal{L}_{\text{align}},
\label{eq:align}
\end{equation}
where $n$ indexes patches and $d$ is a distance (cosine or squared error), and $\lambda=0.5$. $\mathcal{F}_{\text{perception}}$ is the clean image feature extracted by an external perception model such as DINOv2~\citep{dinov2}. iREPA~\citep{irepa} builds on REPA by transferring more of the spatial structure of the
teacher model (the target representation), which further speeds up convergence.
Appendix~\ref{app:irepa} gives the details.

\paragraph{Latent tokenizers.}
Latent diffusion \citep{ldm} trains the denoiser on the compressed latents of a tokenizer
(a VAE) instead of pixels. VAVAE~\citep{lightningdit} aligns these latents with a pretrained
vision encoder, distilling its semantics into the tokenizer so that the latents are easier to
learn. REPA-E~\citep{repae} goes further: besides bringing in these semantics, it enables
joint optimization of the tokenizer and the diffusion model.

\begin{figure}[t]
  \centering

  \includegraphics[width=1\linewidth]{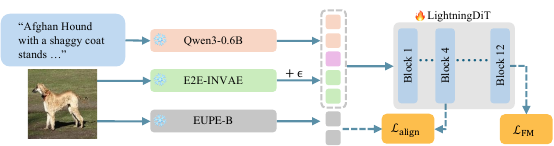}

  \tcbox[colback=black!2!white, colframe=black!25, boxrule=0.4pt, arc=0.8mm,
    boxsep=0pt, left=3pt, right=3pt, top=1.5pt, bottom=1.5pt]{%
    \footnotesize
    \snowemoji\ Frozen\hspace{7pt}\fireemoji\ Trained%
    \hspace{5pt}{\color{black!25}\vrule width 0.4pt}\hspace{5pt}%
    \tokensq{tokText}\ Text embeddings $\mathcal{F}_{\text{text}}$\hspace{7pt}%
    \tokensq{tokTime}\ Time token $\mathcal{F}_{\text{time}}$\hspace{7pt}%
    \tokensq{tokLatent}\ Noisy latents $z_t$\hspace{7pt}%
    \tokensq{tokDino}\ EUPE features $\mathcal{F}_{\text{EUPE}}$}
  \caption{T2I pretraining recipe. Frozen Qwen3~\citep{qwen3}, E2E-INVAE~\citep{repae} and EUPE ViT-B/16~\citep{eupe} produce the text embeddings, the latents (noised with $\epsilon$) and the alignment targets. LightningDiT~\citep{lightningdit} is pretrained with the flow-matching loss~\citep{lipman2022flow} and the iREPA~\citep{irepa} loss on Block 4.} 
  \label{fig:arch}
\end{figure}

\section{The MiniDog resource}
\label{sec:materials}

\textbf{Data.} Pretraining uses 26,000 recaptioned ImageNet~\citep{russakovsky2015imagenet} images of 20 dog breeds. Supervised fine-tuning (SFT) uses 2,000 high-quality images of the same breeds that we generate with FLUX.1-schnell~\citep{flux2024} from text prompts written by GPT-5.5
in the same structure as the real-photo captions. Evaluation uses 500 held-out prompts. All three datasets have been released on \href{https://huggingface.co/datasets/reyhanehesi/minidog-data}{HuggingFace}. 

\textbf{Pipeline} is depicted in Fig.~\ref{fig:arch}, where we adopt the DiffusionBench framework~\citep{diffusionbench}. During pretraining, we pass the image caption through Qwen3-0.6B~\citep{qwen3} and obtain text tokens. The paired image is fed to a VAE and an external vision encoder (\textit{e.g}., DINOv2). Noise is added to the image latent. Timestep and text tokens are concatenated with the noisy latents into one sequence (in-context conditioning~\citep{peebles2023scalable}) and this sequence is fed into the diffusion transformer. The vision perception tokens, on the other hand, are used to align with intermediate diffusion features (REPA loss Eq.~\ref{eq:align}) at Block 4, where our diffusion transformer has 12 blocks in total. The flow matching loss (Eq.~\ref{eq:fm}) is also applied. At the SFT stage, we turn off the REPA loss and only use flow matching loss.

The tokenizer (VAE) is always frozen. Images are encoded into latents which are cached and reused during training: this reduces training time significantly. Unless otherwise stated, we pretrain for 200 epochs ($\sim$20k steps) and SFT for 100 epochs ($\sim$700 steps). Appendix~\ref{app:hparams} lists all hyperparameters.

\textbf{Methods supported}. From architecture point of view, MiniDog supports latent diffusion with a 12-block transformer with 22.06M parameters and in-context conditioning. For VAE, the project supports E2E-INVAE and E2E-VAVAE~\citep{repae}. For REPA, we support original REPA and iREPA. Various vision encoders can be used, including DINOv2, DINOv3~\citep{dinov3}, PE-Spatial~\citep{perceptionencoder}, and EUPE~\citep{eupe}. Our repository also allows for supervised fine-tuning and reward-model-based scoring.




\paragraph{Evaluation metrics supported.} We measure generation quality using gFID~\citep{heusel2017gans}. Specifically, we use all the 26k real pretraining images as the reference set and generate 26k images using the same pretraining prompts. The two sets of images are used for gFID calculation. We measure tokenizer reconstruction using PSNR, SSIM~\citep{ssim}, LPIPS~\citep{lpips} and rFID. To assess the effect of SFT, we use two human-preference models, HPSv2~\citep{hpsv2} and PickScore~\citep{pickapic}. The two scores are averaged over 500 images that the pretrained and the SFT checkpoints (ckpts) each generate from 500 held-out prompts. Inherited from DiffusionBench, our repository supports  online evaluation, where intermediate checkpoints can be evaluated during training. Online evaluation will double the pretraining time and add about an hour to SFT. 
Appendix~\ref{app:metrics} explains each metric in more detail.

\section{Results}
\label{sec:results}

\subsection{Our best recipe}

\textbf{Quality of generated images.} Our best pretraining recipe is to use E2E-INVAE as tokenizer, 128-token captions, iREPA for acceleration with mean squared error loss, and EUPE as the external vision encoder. As shown in Table \ref{tab:prompt}, this method yields gFID = 8.32. We then apply standard SFT. As shown in Table~\ref{tab:pref}, SFT gives us HPSv2 = 0.2427 and PickScore = 19.61.

\textbf{Efficiency.} Using 4$\times$ RTX 3090 GPUs, pretraining for 200 epochs takes 55 minutes, while SFT for 100 epochs takes 15 minutes. If using a single RTX 3090 GPU, pretraining for 200 epochs takes about 3 hours, while SFT for 100 epochs takes 21 minutes. 
 Qualitative results after SFT are shown in Fig.~\ref{fig:misc}. Training the best recipe on a single GPU with the provided configuration (micro-batches of 64 with 4 gradient-accumulation steps, bf16) needs at least 10 GiB of VRAM, or 15 GiB if FID is evaluated during training. Also note that the image and text tokens are precomputed and cached, which takes about 15 minutes on one 3090 GPU. 

\subsection{Method comparisons}
\label{sec:lessons}

\begin{figure}[t]
  \centering
  \begin{minipage}[c]{0.35\linewidth}
    \centering
    \includegraphics[width=\linewidth,trim=0 8 0 0,clip]{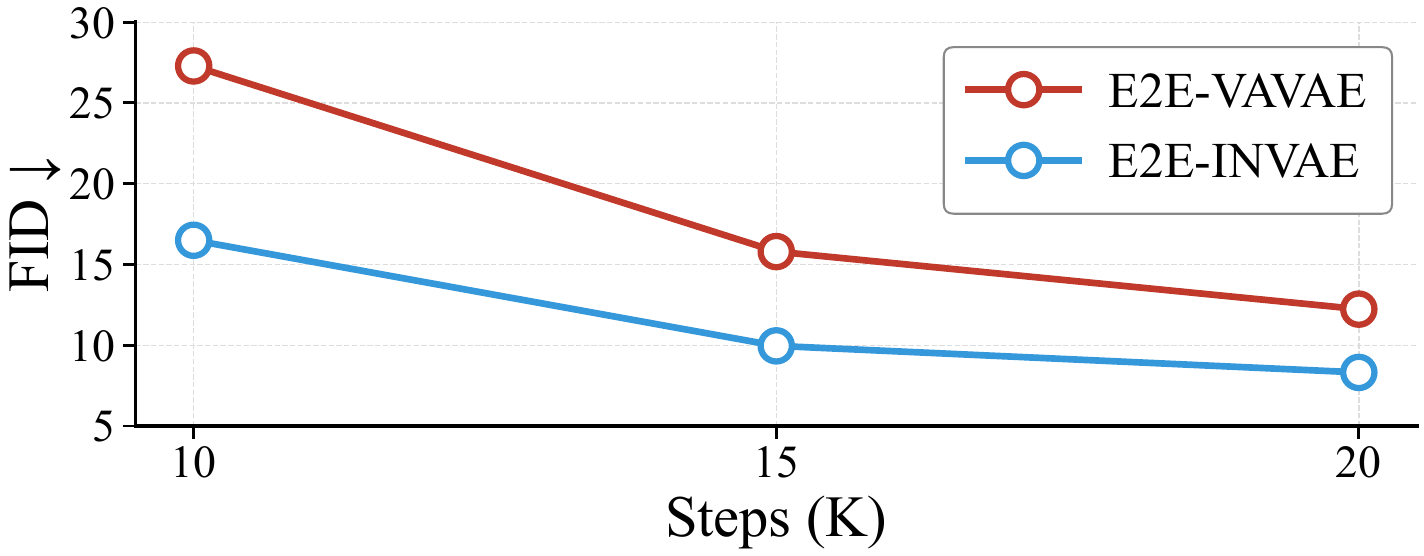}
  \end{minipage}\hfill
  \begin{minipage}[c]{0.34\linewidth}
    \centering
    \scriptsize
    \setlength{\tabcolsep}{2pt}
    \begin{tabular}{@{}lcccc@{}}
      \toprule
                & PSNR$\uparrow$ & SSIM$\uparrow$ & LPIPS$\downarrow$ & rFID$\downarrow$ \\
      \midrule
      E2E-INVAE & \textbf{28.42} & \textbf{0.792} & \textbf{0.039} & \textbf{1.14} \\
      E2E-VAVAE & 27.59          & 0.753          & 0.044          & 1.39          \\
      \bottomrule
    \end{tabular}
  \end{minipage}\hfill
  \begin{minipage}[c]{0.26\linewidth}
    \centering
    \scriptsize
    \setlength{\tabcolsep}{2pt}
    \begin{tabular}{@{}lcc@{}}
      \toprule
      FID$\downarrow$ & 64 tokens & 128 tokens \\
      \midrule
      E2E-INVAE & 8.54 & \textbf{8.32} \\
      \bottomrule
    \end{tabular}
  \end{minipage}

  \begin{minipage}[t]{0.35\linewidth}
    \caption{VAE diffusability comparison. Pretraining gFID with EUPE, iREPA, and 128 tokens.}
    \label{fig:tok_curves}
  \end{minipage}\hfill
  \begin{minipage}[t]{0.34\linewidth}
    \captionof{table}{VAE reconstruction comparison. We report four metrics for the E2E-INVAE and E2E-VAVAE.}
    \label{tab:tokenizers}
  \end{minipage}\hfill
  \begin{minipage}[t]{0.26\linewidth}
    \captionof{table}{Comparing short and long captions FIDs for pretraining stage.}
    \label{tab:prompt}
  \end{minipage}
\end{figure}

\textbf{Comparing E2E-INVAE and E2E-VAVAE}. Both VAEs, trained by the REPA-E method~\citep{repae}, are publicly available. To compare VAEs, both reconstruction and diffusability are important \citep{diffusability}. We first compare their reconstruction performance in Table \ref{tab:tokenizers}. Across the four metrics, E2E-INVAE is consistently better. We then compare how they support diffusion model training in Fig. \ref{fig:tok_curves}. We also observe that E2E-INVAE yields lower gFID scores or better diffusability.

\textbf{Long caption vs. short caption.} For each dog image, we provide a short caption (64 tokens maximum) and a long caption (128 tokens maximum). We compare them in MiniDog using E2E-INVAE in Table \ref{tab:prompt}. For this comparison, iREPA with EUPE encoder and MSE loss is used. Under E2E-INVAE, training with long captions yields lower gFID results. It indicates that long captions allow pretraining to converge faster than short captions.

\begin{figure}[t]
  \centering
  \begin{minipage}[t]{0.40\linewidth}
    \vspace{0pt}
    \centering
    \includegraphics[width=\linewidth,trim=0 8 0 0,clip]{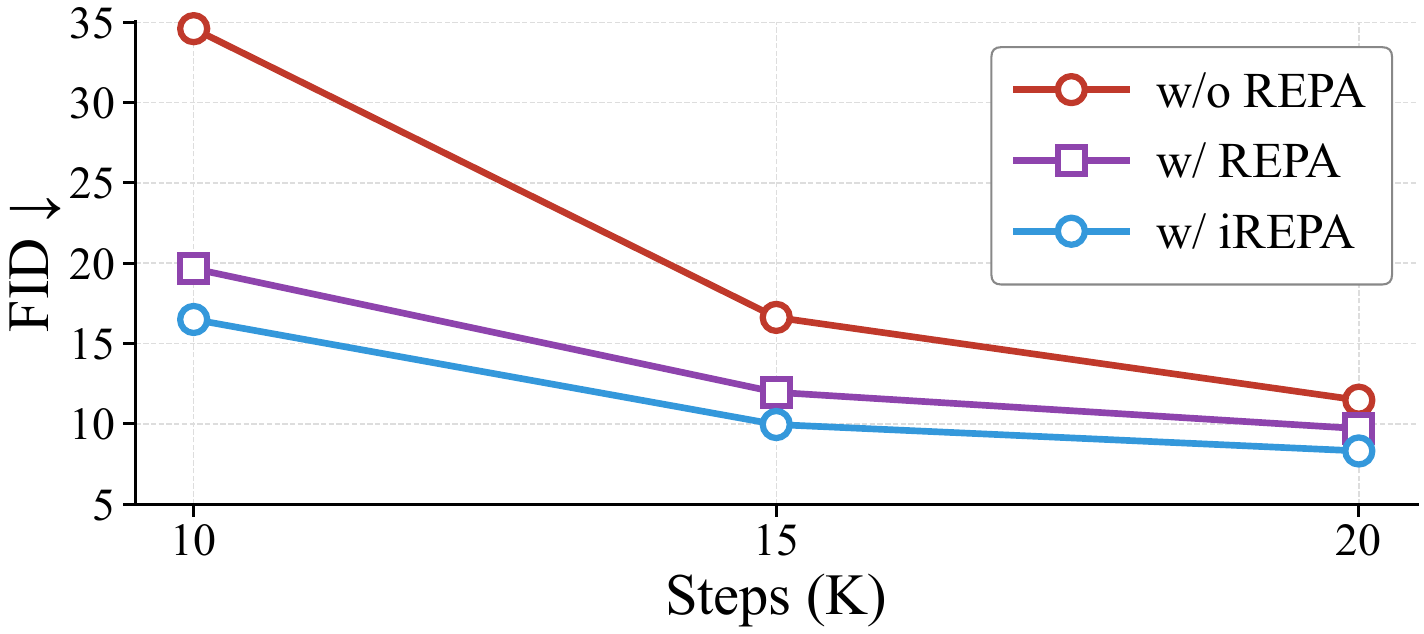}
  \end{minipage}\hfill
  \begin{minipage}[t]{0.57\linewidth}
    \vspace{0pt}
    \centering
    \footnotesize
    \setlength{\tabcolsep}{3pt}
    \begin{tabular}{@{}lcccc@{}}
      \toprule
       & \multicolumn{2}{c}{Squared error} & \multicolumn{2}{c}{Cosine} \\
      \cmidrule(lr){2-3}\cmidrule(lr){4-5}
      Target representation & FID & iREPA FID & FID & iREPA FID \\
      \midrule
      DINOv2 ViT-B/14 & \textbf{8.80} & 8.68          & 9.14  & 8.89 \\
      PE-Spatial B/16 & 9.56          & 9.65          & 9.89  & 9.99 \\
      EUPE ViT-B/16   & 9.71          & \textbf{8.32} & 10.20 & 8.37 \\
      DINOv3 ViT-B/16 & 10.12         & 8.46          & 10.32 & 9.01 \\
      \bottomrule
    \end{tabular}
  \end{minipage}

  \begin{minipage}[t]{0.40\linewidth}
    \caption{Comparing REPA, iREPA, and no REPA. We use E2E-INVAE, EUPE encoder, 128 tokens, and MSE loss.}
    \label{fig:repa_curves}
  \end{minipage}\hfill
  \begin{minipage}[t]{0.57\linewidth}
    \captionof{table}{Comparing external vision encoders, REPA \textit{vs}. iREPA, and MSE \textit{vs}. cosine similarity. We use E2E-INVAE and 128 prompt tokens. $\lambda = 0.5$.}
    \label{tab:teachers}
  \end{minipage}
\end{figure}

\textbf{REPA \textit{vs}. iREPA.} We compare REPA, iREPA and baseline without REPA in Fig. \ref{fig:repa_curves} and Table~\ref{tab:teachers}. We clearly see that REPA accelerates convergence of the baseline. Moreover, when using the EUPE encoder, iREPA converges faster than REPA. While REPA is slightly better under the PE-Spatial encoder (Table~\ref{tab:teachers}), the overall best performance is still achieved with iREPA (with EUPE). In addition, from Table~\ref{tab:teachers}, using the MSE loss is superior to the cosine similarity loss for REPA and iREPA.

\textbf{Comparing different vision encoders.} In Table~\ref{tab:teachers}, we observe that the best encoder is DINOv2 for REPA and EUPE for iREPA. The EUPE encoder~\citep{eupe} is trained by distilling from DINOv3, PE-Core, and PE-Lang. Its spatially coherent dense features may explain its advantages for iREPA.

\begin{wraptable}{r}{0.35\textwidth}
  \centering
  \footnotesize
  \setlength{\tabcolsep}{3.5pt}
  \begin{tabular}{lcc}
    \toprule
               & HPSv2  & PickScore \\
    \midrule
    Pretrained & 0.1625 & 18.42  \\
    SFT        & \textbf{0.2427} & \textbf{19.61} \\
    \bottomrule
  \end{tabular}
  \normalsize
  \caption{Human-preference scores of the pretrained and SFT ckpts.}
  \label{tab:pref}
\end{wraptable}
\textbf{Benefit of supervised fine-tuning.} In Table~\ref{tab:pref}, we compare generation quality of our pretrained model and SFT model. We use two pretrained reward models for scoring, which generally reflect how much people would like an image given its prompt. It is obvious that the HPSv2 and PickScore of the SFT model are consistently higher than those of the pretrained model. It is mainly because of the high quality SFT data. Note that FID (not shown) of the SFT model is higher than the pretrained model, because the high-quality SFT data pulls the model away from the pretraining distribution. Fig.~\ref{fig:misc} presents some cherry-picked images.

\section{Conclusion}

MiniDog does not build a comprehensive T2I model. Instead, it is a minimal project that focuses on dog image generation for the purpose of democratizing T2I training. Under our configuration, pretraining only takes about an hour on four 3090 GPUs and yields an FID of 8.32 at the resolution of $256\times256$. Users can spend another 15 minutes on supervised fine-tuning, which improves image quality judged by pretrained reward models. The entire project including data, models, and code has been publicly open-sourced, with the hope to allow researchers to get started and evaluate their ideas.

\renewcommand{\refname}{References}
\vspace{-1em}
{
\setlength{\bibsep}{2pt}
\bibliographystyle{unsrtnat}
\bibliography{references}

@article{lipman2022flow,
  title={Flow matching for generative modeling},
  author={Lipman, Yaron and others},
  journal={ICLR},
  year={2023}
}

@inproceedings{peebles2023scalable,
  title={Scalable diffusion models with transformers},
  author={Peebles, William and Xie, Saining},
  booktitle={ICCV},
  year={2023}
}

@article{ho2022classifier,
  title={Classifier-free diffusion guidance},
  author={Ho, Jonathan and Salimans, Tim},
  journal={arXiv:2207.12598},
  year={2022}
}

@inproceedings{repa,
  title={Representation alignment for generation: Training diffusion transformers is easier than you think},
  author={Yu, Sihyun and others},
  booktitle={ICLR},
  year={2025}
}

@article{dinov2,
  title={{DINOv2}: Learning robust visual features without supervision},
  author={Oquab, Maxime and others},
  journal={TMLR},
  year={2024}
}

@article{pickapic,
  title={{Pick-a-Pic}: An open dataset of user preferences for text-to-image generation},
  author={Kirstain, Yuval and others},
  journal={NeurIPS},
  year={2023}
}

@article{hpsv2,
  title={Human preference score v2: A solid benchmark for evaluating human preferences of text-to-image synthesis},
  author={Wu, Xiaoshi and others},
  journal={arXiv:2306.09341},
  year={2023}
}

@article{qwen3,
  title={Qwen3 technical report},
  author={Yang, An and others},
  journal={arXiv:2505.09388},
  year={2025}
}

@inproceedings{repae,
  title={{REPA-E}: Unlocking {VAE} for End-to-End Tuning of Latent Diffusion Transformers},
  author={Leng, Xingjian and others},
  booktitle={ICCV},
  year={2025}
}

@inproceedings{lightningdit,
  title={Reconstruction vs. generation: Taming optimization dilemma in latent diffusion models},
  author={Yao, Jingfeng and others},
  booktitle={CVPR},
  year={2025}
}

@article{diffusionbench,
  title={{DiffusionBench}: On Holistic Evaluation of Diffusion Transformers},
  author={Leng, Xingjian and others},
  journal={arXiv:2606.24888},
  year={2026}
}

@inproceedings{rectifiedflow,
  title={Flow straight and fast: Learning to generate and transfer data with rectified flow},
  author={Liu, Xingchao and others},
  booktitle={ICLR},
  year={2023}
}

@article{russakovsky2015imagenet,
  title={{ImageNet} large scale visual recognition challenge},
  author={Russakovsky, Olga and others},
  journal={IJCV},
  year={2015}
}

@misc{flux2024,
  author={{Black Forest Labs}},
  title={{FLUX}},
  year={2024},
  howpublished={\url{https://github.com/black-forest-labs/flux}}
}

@article{heusel2017gans,
  title={{GAN}s trained by a two time-scale update rule converge to a local {N}ash equilibrium},
  author={Heusel, Martin and others},
  journal={NeurIPS},
  year={2017}
}

@article{rae,
  title={Diffusion transformers with representation autoencoders},
  author={Zheng, Boyang and others},
  journal={ICLR},
  year={2026}
}

@article{dinov3,
  title   = {{DINOv3}},
  author  = {Sim{\'e}oni, Oriane and Vo, Huy V. and others},
  journal = {TMLR},
  year    = {2026},
}

@article{eupe,
  title   = {Efficient Universal Perception Encoder},
  author  = {Zhu, Chenchen and Suri, Saksham and others},
  journal = {arXiv:2603.22387},
  year    = {2026},
}

@article{irepa,
  title   = {What Matters for Representation Alignment: Global Information or Spatial Structure?},
  author  = {Singh, Jaskirat and Leng, Xingjian and others},
  journal = {ICLR},
  year    = {2026},
}

@inproceedings{perceptionencoder,
  title     = {Perception Encoder: The Best Visual Embeddings Are Not at the Output of the Network},
  author    = {Bolya, Daniel and Huang, Po-Yao and others},
  booktitle = {NeurIPS},
  year      = {2025},
  }

@inproceedings{ldm,
  title     = {High-Resolution Image Synthesis with Latent Diffusion Models},
  author    = {Robin Rombach and Andreas Blattmann and Dominik Lorenz and
               Patrick Esser and Bj{\"o}rn Ommer},
  booktitle = {CVPR},
  year      = {2022},
}

@inproceedings{ddpm,
  title     = {Denoising Diffusion Probabilistic Models},
  author    = {Jonathan Ho and Ajay Jain and Pieter Abbeel},
  booktitle = {NeurIPS},
  year      = {2020},
}

@article{ssim,
  title   = {Image Quality Assessment: From Error Visibility to Structural Similarity},
  author  = {Zhou Wang and Alan C. Bovik and Hamid R. Sheikh and Eero P. Simoncelli},
  journal = {IEEE Transactions on Image Processing},
  volume  = {13},
  number  = {4},
  pages   = {600--612},
  year    = {2004},
}

@inproceedings{lpips,
  title     = {The Unreasonable Effectiveness of Deep Features as a Perceptual Metric},
  author    = {Richard Zhang and Phillip Isola and Alexei A. Efros and Eli Shechtman and Oliver Wang},
  booktitle = {CVPR},
  year      = {2018},
}

@inproceedings{diffusability,
  title     = {Improving the Diffusability of Autoencoders},
  author    = {Skorokhodov, Ivan and Girish, Sharath and Hu, Benran and
               Menapace, Willi and Li, Yanyu and Abdal, Rameen and
               Tulyakov, Sergey and Siarohin, Aliaksandr},
  booktitle = {ICML},
  year      = {2025}
}

@inproceedings{vit,
  title={An Image is Worth 16x16 Words: Transformers for Image Recognition at Scale},
  author={Dosovitskiy, Alexey and others}, booktitle={ICLR}, year={2021}}

@inproceedings{unet,
  title={{U-Net}: Convolutional Networks for Biomedical Image Segmentation},
  author={Ronneberger, Olaf and others}, booktitle={MICCAI}, year={2015}}

@inproceedings{sd3,
  title={Scaling Rectified Flow Transformers for High-Resolution Image Synthesis},
  author={Esser, Patrick and others}, booktitle={ICML}, year={2024}}

@inproceedings{inceptionv3,
  title={Rethinking the {Inception} Architecture for Computer Vision},
  author={Szegedy, Christian and others}, booktitle={CVPR}, year={2016}}

@inproceedings{clip,
  title={Learning Transferable Visual Models From Natural Language Supervision},
  author={Radford, Alec and others}, booktitle={ICML}, year={2021}}

@misc{torchfidelity,
  title={High-fidelity performance metrics for generative models in {PyTorch}},
  author={Obukhov, Anton and others}, year={2020},
  howpublished={\url{https://github.com/toshas/torch-fidelity}}}

@inproceedings{rmsnorm,
  title={Root Mean Square Layer Normalization},
  author={Zhang, Biao and Sennrich, Rico}, booktitle={NeurIPS}, year={2019}}

@article{swiglu,
  title={{GLU} Variants Improve Transformer},
  author={Shazeer, Noam}, journal={arXiv:2002.05202}, year={2020}}

@article{rope,
  title={{RoFormer}: Enhanced Transformer with Rotary Position Embedding},
  author={Su, Jianlin and others}, journal={Neurocomputing}, year={2024}}

}

\appendix

\section{iREPA}
\label{app:irepa}

iREPA~\citep{irepa} keeps the REPA loss of Eq.~\eqref{eq:align} but makes two changes that
transfer more of the spatial structure of the target representation. The iREPA runs done during the
comparison of target representations (Table~\ref{tab:teachers}) make both changes and keep
everything else fixed.

\paragraph{Convolutional projector.} REPA's MLP projector $g_\phi$(a single linear layer in our
REPA runs) is replaced by a $3\times3$
convolution (padding 1) over the $16\times16$ grid of hidden states, which preserves the local
spatial relations between neighbouring patches.

\paragraph{Spatial normalization.} The target features are normalized across the patch tokens
of each image before alignment,
\begin{equation*}
\tilde{x}_n = \frac{x_n - \gamma\,\bar{x}}{\sigma + \epsilon},
\end{equation*}
where $x_n$ is the feature of patch $n$, $\bar{x}$ and $\sigma$ are the mean and standard
deviation over the patches of the image, $\gamma = 0.7$, and $\epsilon = 10^{-6}$. Subtracting the mean removes
most of the global component that all patches share, which increases the contrast between patches
and so strengthens the spatial signal that alignment transfers to the backbone.
Table~\ref{tab:hparams} in Appendix~\ref{app:hparams} lists these settings.

\section{Evaluation metrics}
\label{app:metrics}

This appendix explains the metrics of Section~\ref{sec:materials}. For each, an arrow in the
tables marks whether lower ($\downarrow$) or higher ($\uparrow$) is better.

\paragraph{FID $\downarrow$.} The Fr\'echet Inception Distance~\citep{heusel2017gans} compares
two sets of images through the features of a pretrained InceptionV3~\citep{inceptionv3} network. It fits a
Gaussian to the features of each set, with means $\mu_r,\mu_g$ and covariances
$\Sigma_r,\Sigma_g$, and measures the distance between the two Gaussians:
\begin{equation*}
\mathrm{FID} = \lVert \mu_r - \mu_g \rVert^2
  + \mathrm{Tr}\bigl(\Sigma_r + \Sigma_g - 2(\Sigma_r\Sigma_g)^{1/2}\bigr).
\end{equation*}
It captures both the quality and the diversity of the generated images. We compare generated
images against the 26k pretraining images.

\paragraph{rFID $\downarrow$.} The same FID, computed between the pretraining images and their
reconstructions after encoding and decoding with the tokenizer. It measures how much the
tokenizer loses, and because it uses the same Inception reference as our FID, it can be read on
the same scale.

\paragraph{PSNR $\uparrow$.} The peak signal-to-noise ratio,
$10\log_{10}(\mathrm{MAX}^2/\mathrm{MSE})$ in dB, where MSE is the mean squared pixel error
between an image and its reconstruction and MAX is the largest possible pixel value. It
measures pixel-level fidelity.

\paragraph{SSIM $\uparrow$.} The structural similarity index~\citep{ssim} compares local
luminance, contrast and structure between an image and its reconstruction. It equals 1 for
identical images.

\paragraph{LPIPS $\downarrow$.} The learned perceptual image patch similarity~\citep{lpips}
measures the distance between the deep-network features of two images, calibrated so that it
agrees with human judgments of perceptual similarity.

\paragraph{HPSv2 and PickScore $\uparrow$.} Both are CLIP-based~\citep{clip} models fine-tuned on human
choices between pairs of generated images: HPSv2~\citep{hpsv2} on the Human Preference Dataset
v2, and PickScore~\citep{pickapic} on the Pick-a-Pic dataset. Given a prompt and an image, each
predicts how much people would prefer the image. We use them only for evaluation, never as a
training signal, and report the average score over the 500 images, one per held-out prompt, that
each checkpoint (pretrained and SFT) generates.

\section{Hyperparameters and more implementation details}
\label{app:hparams}

\begin{table}[t]
  \centering
  \footnotesize
  \setlength{\tabcolsep}{10pt}
  \begin{tabular}{lcc}
    \toprule
                               & Pretraining                       & SFT \\
    \midrule
    \multicolumn{3}{@{}l}{\emph{Architecture}} \\
    Backbone                   & \multicolumn{2}{c}{LightningDiT} \\
    Input dim.                 & \multicolumn{2}{c}{$16\times16\times32$} \\
    Patch size                 & \multicolumn{2}{c}{1} \\
    Num.\ layers               & \multicolumn{2}{c}{12} \\
    Hidden dim.                & \multicolumn{2}{c}{384} \\
    Num.\ heads                & \multicolumn{2}{c}{6} \\
    Timestep / text tokens     & \multicolumn{2}{c}{4 / 128} \\
    Text encoder               & \multicolumn{2}{c}{Qwen3-0.6B (frozen)} \\
    Tokenizer                  & \multicolumn{2}{c}{E2E-INVAE (frozen)} \\
    \midrule
    \multicolumn{3}{@{}l}{\emph{Alignment --- final recipe}} \\
    Target representation      & EUPE ViT-B/16                     & -- \\
    Method                     & iREPA                             & -- \\
    Distance $d$               & mean squared error                & -- \\
    $\lambda$                  & 0.5                               & -- \\
    Alignment depth            & 4                                 & -- \\
    Projector $g_\phi$         & $3\times3$ conv, $384\to768$, padding 1 & -- \\
    Spatial norm.\ $\gamma$, $\epsilon$ & 0.7, $10^{-6}$           & -- \\
    \midrule
    \multicolumn{3}{@{}l}{\emph{Alignment --- compared in Table~\ref{tab:teachers}}} \\
    Target representations     & DINOv2, DINOv3, PE-Spatial, EUPE  & -- \\
    Methods                    & REPA (linear, $384\to768$), iREPA & -- \\
    Distances $d$              & mean squared error, cosine        & -- \\
    \midrule
    \multicolumn{3}{@{}l}{\emph{Optimization}} \\
    Training data              & 26k images                        & 2k images \\
    Epochs                     & 200 ($\sim$20k steps)             & 100 \\
    Batch size                 & 256                               & 256 \\
    Optimizer                  & AdamW                             & AdamW \\
    Learning rate              & $10^{-4}$, then linear decay      & $5\times10^{-5}$, constant \\
    $(\beta_1, \beta_2)$       & $(0.9, 0.999)$                    & $(0.9, 0.999)$ \\
    Weight decay               & 0                                 & 0 \\
    Gradient clipping          & 1.0                               & 1.0 \\
    EMA decay                  & 0.9995                            & 0.995 \\
    Condition dropout          & 0.1                               & 0.1 \\
    Precision                  & bf16                              & bf16 \\
    \midrule
    \multicolumn{3}{@{}l}{\emph{Interpolants and sampling}} \\
    $\alpha_t$, $\sigma_t$     & \multicolumn{2}{c}{$1-t$, $t$} \\
    Training objective         & \multicolumn{2}{c}{v-prediction} \\
    Timestep sampling          & \multicolumn{2}{c}{logit-normal$(0,1)$, shift $\sqrt{2}$} \\
    Sampler                    & \multicolumn{2}{c}{Euler (ODE), 50 steps} \\
    Guidance (CFG scale)       & \multicolumn{2}{c}{6.0} \\
    \bottomrule
  \end{tabular}

  \normalsize
  \caption{Hyperparameters of the final recipe (\texttt{configs/\allowbreak pretrain\_e2e-invae\_128tok\_mse\_irepa\_eupe.yaml} and
  \texttt{configs/\allowbreak sft\_e2e-invae\_128tok\_mse\_norepa.yaml}).}
  \label{tab:hparams}
\end{table}

\paragraph{Architecture.} We use LightningDiT~\citep{lightningdit} with RMSNorm~\citep{rmsnorm},
query-key normalization, SwiGLU feed-forward layers~\citep{swiglu}, and 2D rotary position embeddings~\citep{rope}.
Conditioning is in-context: the timestep is embedded into 4 tokens with Gaussian Fourier
features, the text embedding $\mathcal{F}_{\text{text}}$ (the last hidden layer of Qwen3-0.6B,
padded to 128 tokens with an attention mask) is projected to the hidden size, and both are
concatenated with the 256 latent tokens.

\paragraph{Cached inputs.} Every frozen model runs only once, before training. We resize and
center-crop each image to $256\times256$ and cache its E2E-INVAE latent, its text embedding, and,
for alignment, the patch features of the target representation (DINOv2, or EUPE in the final
recipe), so training updates only the DiT. We use no data
augmentation. The 64-token ablations use a second, shorter caption (about 30--40 words) for each image,
so that the caption fits within 64 tokens.

\paragraph{Optimization.} We use AdamW with $(\beta_1,\beta_2) = (0.9, 0.999)$, no weight decay,
gradient clipping at 1.0, and bf16 mixed precision, and we evaluate an exponential moving average
(EMA) of the weights. Pretraining keeps the learning rate at $10^{-4}$ for 100 epochs and then
decays it linearly; the schedule would reach $10^{-5}$ at epoch 600, so at the end of training
(epoch 200) the learning rate is $8.2\times10^{-5}$. SFT starts from the pretrained weights with a
fresh optimizer and a constant learning rate of $5\times10^{-5}$. In both stages, the text
condition is dropped with probability 0.1 so that classifier-free guidance can be used at sampling
time.


\paragraph{REPA.} A single linear layer $g_\phi$ maps the 384-dimensional hidden states of
block 4 to the 768-dimensional patch features of the target representation, and
$\mathcal{L}_{\text{align}}$ is their mean squared error (or $1-\cos$ per patch in the cosine
runs of Table~\ref{tab:teachers}), weighted by $\lambda = 0.5$. All images are $256\times256$.
DINOv3, PE-Spatial and EUPE, which use $16\times16$ patches, take them at that size; DINOv2
ViT-B/14, used without register tokens, takes them resized to $224\times224$, with its
positional embeddings resampled to a $16\times16$ grid. Every target representation therefore
gives 256 patch tokens, which match the 256 latent tokens one to one.

\paragraph{Target representations.} The target representations of Table~\ref{tab:teachers} come
from DINOv2 ViT-B/14, DINOv3 ViT-B/16, PE-Spatial B/16, and EUPE ViT-B/16, all with
768-dimensional patch features.
DINOv2 and DINOv3 are used as released, with their final LayerNorm and its learned affine
parameters. We renormalise the other two, so that every encoder ends in a LayerNorm: PE-Spatial
has no output normalisation, so we add a parameter-free LayerNorm, and for EUPE we remove the
learned affine parameters of its final LayerNorm.

\paragraph{Timesteps and sampling.} Training timesteps follow a logit-normal distribution,
shifted towards noise by $t \mapsto s\,t/\bigl(1+(s-1)\,t\bigr)$ with the dimension-dependent
factor $s = \sqrt{8192/4096} = \sqrt{2}$, where $8192 = 16\cdot16\cdot32$ is the latent
dimension~\citep{rae}. Sampling integrates the learned velocity with 50 Euler steps from $t=1$ to
$t=0$ on a time grid with the same shift, and applies classifier-free
guidance~\citep{ho2022classifier} with scale 6.0 at every step.

\paragraph{Evaluation.} FID uses the InceptionV3 network of torch-fidelity~\citep{torchfidelity}, one generated image
per caption of the 26k pretraining set, and reference statistics of the 26k real images. HPSv2
and PickScore use their officially released checkpoints.


\paragraph{Computing resources.} All runs except the single-GPU timings in Section 4.1 use 4 of the 8 NVIDIA RTX 3090 GPUs (24\,GB, 350\,W)
in a single server, connected over PCIe~4.0 without NVLink, with two AMD EPYC 7343 CPUs
(32 cores in total), 256\,GB of RAM, and the cached inputs on a local NVMe SSD. The software
stack is Ubuntu~24.04, NVIDIA driver 590.44, Python~3.13, PyTorch~2.10.0 with CUDA~12.8,
cuDNN~9.10, NCCL~2.27, and Triton~3.6; \texttt{pyproject.toml} and \texttt{uv.lock} in the
repository pin every package version. Training uses PyTorch DistributedDataParallel over the
4 GPUs (64 images per GPU).

\end{document}